\def\year{2019}\relax
\documentclass[letterpaper]{article} 
\usepackage{aaai19}  
\usepackage{times}  
\usepackage{helvet}  
\usepackage{courier}  
\usepackage{url}  
\usepackage{graphicx}  
\usepackage{booktabs}       
\usepackage{algorithm}
\usepackage{algorithmic}
\usepackage{amsmath,amstext,amssymb}
\newtheorem{theorem}{Theorem}

\newcommand{\R}{{\mathbb{R}}}
\newcommand{\Z}{{\mathbb{Z}}}
\newcommand{\be}{\begin{eqnarray}}
\newcommand{\ee}{\end{eqnarray}}
\newcommand{\beq}{\begin{equation}\begin{aligned}}
\newcommand{\eeq}{\end{aligned}\end{equation}}
\newcommand{\beqn}{\begin{equation*}\begin{aligned}}
\newcommand{\eeqn}{\end{aligned}\end{equation*}}
\newcommand{\ben}{\begin{eqnarray*}}
\newcommand{\een}{\end{eqnarray*}}

\begin{document}
%
\title{Task-Driven Common Representation Learning\\via Bridge Neural Network}
 \author{  Yao Xu\textsuperscript{1,2}, Xueshuang Xiang\textsuperscript{1,2,}\thanks{Corresponding author. }, Meiyu Huang\textsuperscript{1} \\
 	\textsuperscript{1}Qian Xuesen Laboratory of Space 
 	Technology, China Academy of 
 	Space Technology, Beijing, 100190\\
 	\textsuperscript{2}School of Aerospace Science 
 	and Technology, Xidian University, 
 	Xian, 710071 \\
 	\texttt{\{xuyao, 
 	xiangxueshuang, huangmeiyu\}@qxslab.cn}\\
 }
\maketitle
\begin{abstract}
	This paper introduces a novel deep learning based method, named bridge neural network (BNN) to dig the potential relationship between two given data sources task by task. The proposed approach employs two convolutional neural networks that project the two data sources into a feature space to learn the desired common representation required by the specific task. The training objective with artificial negative samples is introduced with the ability of mini-batch training and it's asymptotically equivalent to maximizing the total correlation of the two data sources, which is verified by the theoretical analysis. The experiments on the tasks, including pair matching, canonical correlation analysis, transfer learning, and reconstruction demonstrate the state-of-the-art performance of BNN, which may provide new insights into the aspect of common representation learning.
\end{abstract}

\section{Introduction}
\begin{figure}[htb]
	\begin{center}
		\includegraphics[width=0.42\textwidth]{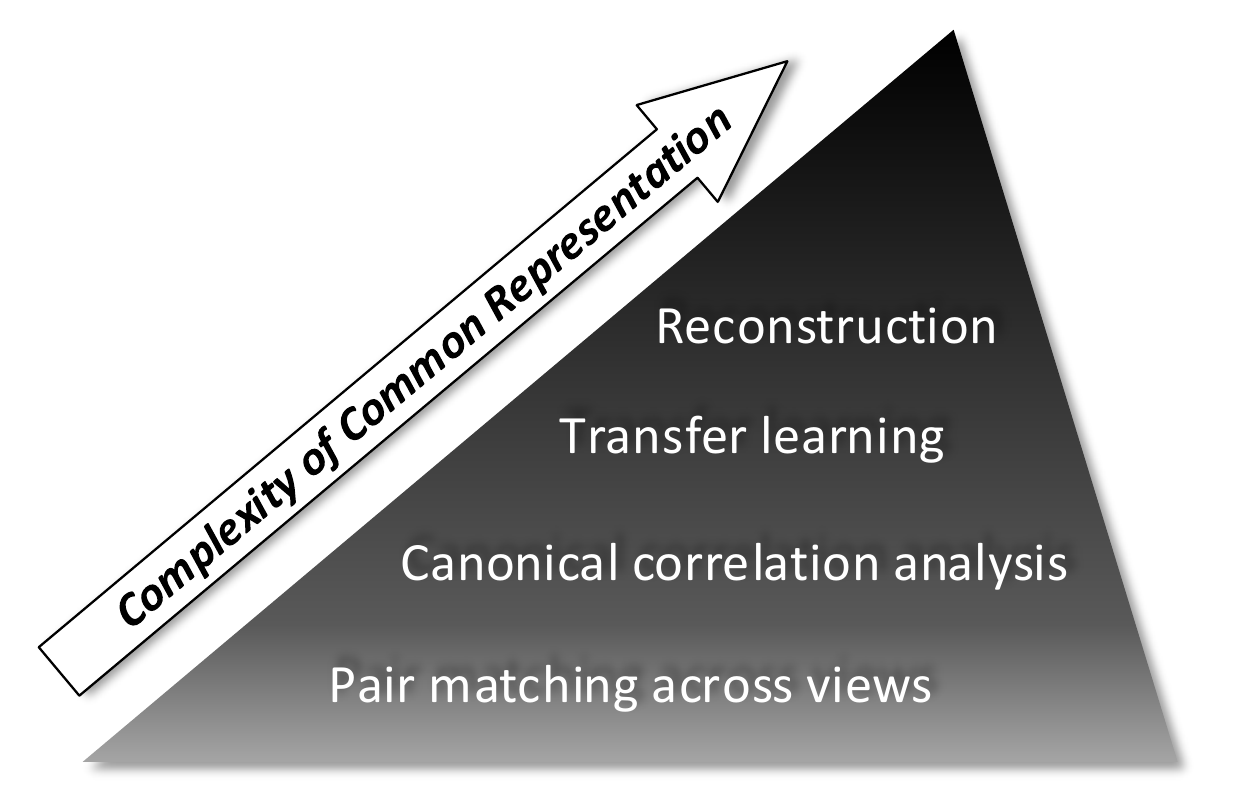}
	\end{center}
	\caption{The complexity of common representations varying from different tasks.
		A task on the top of the triangle is more complex than a bottom one. }\label{fig1a}
\end{figure}

In the real world, a potential relationship always exists between two sets of data, which can be either from multi-views of one data source, e.g., two voices of songs, audio and subtitles of movies, or from two different data sources, e.g., faces of couples, objects with the same labels. One idea of mining the potential relationship of given data pairs is learning a common representation of them, which has achieved lots of interests~\cite{ngiam2011multimodal,eisenschtat2017linking,andrew2013deep,chandar2016correlational}. These methods tried to build a universal model motivated by more than one task, including (i) pair matching across views, (ii) canonical correlation analysis (CCA)~\cite{hotelling1936relations}, (iii) transfer learning and (iv) reconstruction of a missing view. However, different task imposes different complexity levels of common representations, as shown in Figure \ref{fig1a}.

To clarify Figure~\ref{fig1a}, the complexity level of common representations for each task will be described briefly, from simple to complex. (i) For the task of pair matching across views, the common representations are just some potential information which can build a connection between data pairs. This connection can be just a few similarities between pairs, which can be fairly simple. (ii) For the task of CCA, the common representations are the projections of two data sources and it is supposed to maximize the linear correlation between them. CCA methods are always developed as a strategy to build common representations, which makes it an intermediate task rather than a final application objective. (iii) For the task of transfer learning, the common representations need to be features which are useful for a certain learning objective, such as classification. The relationship of those features can be more complex than the linear correlation. (iv) Reconstruction of a missing view requires the common representations to be an encoder, which should contain the information of the views as whole as possible.

\begin{figure}[htb]
	\begin{center}
		\includegraphics[width=0.42\textwidth]{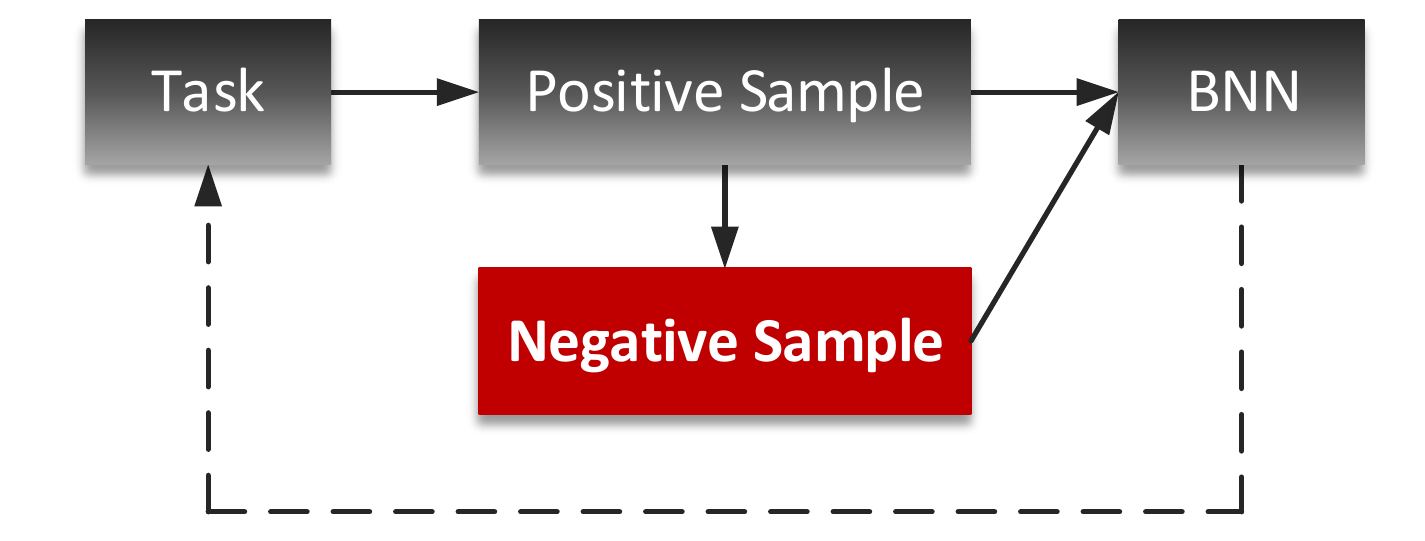}
	\end{center}
	\caption{A framework of using Bridge Neural Network (BNN) to handle a task, e.g. as that shown in Figure \ref{fig1a}. }\label{fig1b}
\end{figure}

Because of the diversity of different tasks, different tasks impose different levels of complexity of the common representations. Motivated by this understanding, we propose a novel deep learning based method, named bridge neural network (BNN) to learn common representations by mining the potential relationship of the specified data sources according to a given task, as shown in Figure \ref{fig1b}. Given a task, we first construct the correlated data source, named positive samples and build the negative samples according to the positive samples. Then both positive samples and negative samples are used to train a BNN to handle the given task.

\begin{figure*}[htb]
	\begin{center}
			\includegraphics[width=0.98\textwidth]{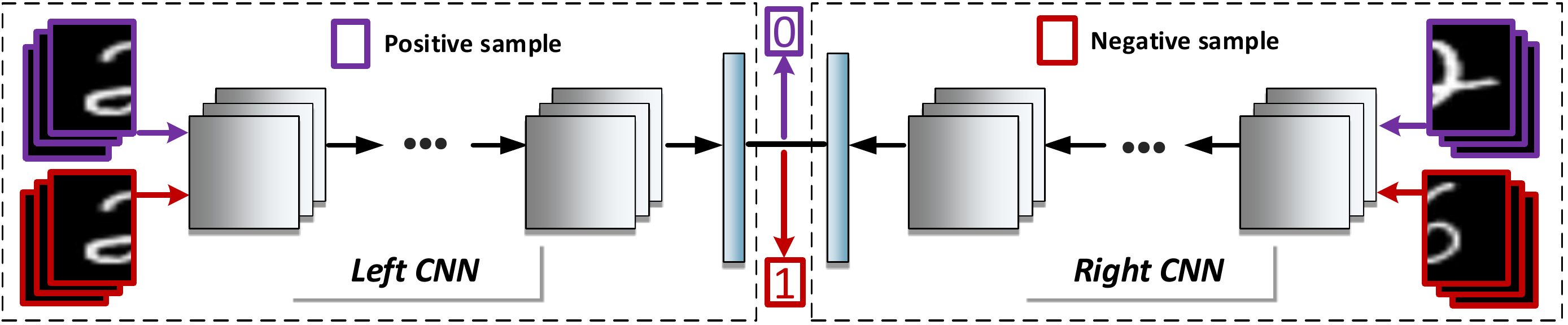}
	\end{center}
	\caption{A schematic of bridge neural network, which employs two convolutional neural networks that project two given data sources into a common feature space. The Euclidean loss of the two output layers is supposed to close to $0$ or $1$ for positive samples or negative samples respectively.
		For the task of reconstruction of a missing view, transposed convolution networks should be introduced to make the common representation an encoder, see Figure \ref{reconmodel}. }\label{fig2}
\end{figure*}

As that shown in Figure \ref{fig2}, BNN employs two convolutional neural networks that project the two data sources into a common feature space and use the Euclidean loss to determine whether two given data sources have a potential relationship, i.e. positive sample or not, i.e. negative sample. Thus, the problem of mining the potential relationship can be transferred to a binary classification problem by introducing artificial negative samples. The main contributions of the proposed BNN in this paper are:
\begin{itemize}
	\item First propose a task-driven framework to learn common representations by mining the potential relationship of given data pairs, which is specified task by task;
	\item A novel optimization problem, i.e. training objective using artificial negative samples is introduced, and it's asymptotically equivalent to maximization of the total correlation;
	\item BNN with lightweight convolution layers can be trained using Gradient Descent based optimization methods, making it more scalable for dealing with large high dimensional data.
\end{itemize}

\section{Related work}
Canonical Correlation Analysis (CCA)~\cite{hotelling1936relations} is often used to build common representations. It is also a general procedure for investigating the relationships between two sets of variables by computing a linear projection for data pairs into a common space which can maximize their linear correlation. 
It plays a significant role in many fields including biology and neurology~\cite{hardoon2007unsupervised}, natural language processing~\cite{dhillon2011multi}, speech processing~\cite{arora2013multi} and computer vision tasks, e.g., action recognition~\cite{kim2007tensor}, linking text and image~\cite{eisenschtat2017linking}. CCA is also a basic method in multi-view learning, see~\cite{Xu2013A,Zhao2017Multi} for details. 

However, traditional CCA method~\cite{hotelling1936relations} and its derivatives, such as regularized CCA~\cite{vinod1976canonical}, Nonparametric Canonical Correlation Analysis (NCCA)~\cite{michaeli2016nonparametric}, Randomized Canonical Correlation Analysis (RCCA)~\cite{mineiro2014randomized} and Kernel Canonical Correlation Analysis (KCCA)~\cite{akaho2006kernel,hardoon2004canonical,bach2002kernel,melzer2001nonlinear}, do not scale well with the size of the dataset and the representations. A number of researches were therefore proposed to overcome this drawback. Deep Canonical Correlation Analysis (DCCA)~\cite{andrew2013deep} and its improved versions in image and text matching~\cite{yan2015deep,wang2016learning} are one of those which introduced deep learning method. Based on DCCA, later works such as Correlation Neural Network(CorrNet)~\cite{chandar2016correlational} and Deep Canonically Correlated Autoencoders (DCCAE)~\cite{wang2015deep}, brought in Multimodal Autoencoder (MAE)~\cite{ngiam2011multimodal} to extend the task of reconstruction of views. The introducing of MAE also improves the performance of CCA because it aims to capture a more meaningful common representation by adding an optimization objective minimizing the distance between the original input and the decoded output of each view. Along this line, 2-Way Nets (2WayNet)~\cite{eisenschtat2017linking} gave a different approach by replacing the optimization objective into the Euclidean loss. In summary, these methods achieved the state-of-the-art performance, but still have some problems, such as the scalability, high inference time with fully connected layers and most of these methods were to build a universal model motivated by more than one task. 

In order to fix these problems, this paper proposes bridge neural network (BNN) with lightweight convolution layers to learn common representations by mining the potential relationship of the specified data sources according to a given task. The most related work is DCCA~\cite{andrew2013deep}, 2WayNet~\cite{eisenschtat2017linking} and Siamese Network ~\cite{chopra2005learning}, which is a similar approach of using two networks to learn the similarity between two inputs. Differences between BNN and each of them will be described in next section.

\section{Bridge Neural Network}
\label{BNN}
This section contains a detailed description of our proposed model. It is termed as Bridge Neural Network (BNN) because it acts as a bridge connecting two sets of variables by projecting them to a common feature subspace. Instead of telling the neural network to use CCA to investigate the relationship between two sets of variables~\cite{andrew2013deep}, we try to tell it which pairs of views are relevant and which are not, hoping the network to find a rule to analyze the potential relationship of the given data sources. The goal of BNN is to learn a common representation by mining the potential relationship of the specified data sources according to a given computer vision task.
\subsection{Data sources}
As we discussed in Figure \ref{fig1a}, different tasks impose different complexity levels of common representations. Given a task, we should first construct the data pairs, which can learn common representations of the corresponding complexity level.  
For example, suppose we try to build a model to match items across views, we can set the data pairs as the sets of two data sources, denoted by $\{X_1, X_2\} \subset \R^{n_1} \times \R^{n_2}$, $X_1 = \{x_1^i\}_{i=1}^{N}$, $X_2 = \{x_2^i\}_{i=1}^{N}$.
And the i-{th} component  $x_1^i\in X_1$, $x_2^i\in X_2$ match each other. Here we define $S_p = \{x_1^i, x_2^i\}$ as the positive sample set and $S_n = \{x_1^i, x_2^j\}, i\neq j$ as the negative sample set.
For the matching task, since the i-{th} components $x_1^i\in X_1$, $x_2^i\in X_2$ are supposed to match each other, a heuristic idea is the i-{th} component  $x_1^i\in X_1$ should not match the j-{th} component  $x_2^j\in X_2$, $i \neq j$. Actually, in practice, we may have two components $x_1^i, x_1^j\in X_1$ very close to each other. If $x_1^i$ matches $x_2^i$ and $x_1^j$ matches $x_2^j$, we may also have $x_1^i$ matches $x_2^j$. Thus in the training process, the negative samples used to train BNN model are randomly selected from $S_n$, where we also take account the situation that the size of $S_n$ is $N(N-1)$, that is almost square of the size of positive sample. The details of constructing data pairs according to different computer vision task can be found in the Experiments.

\subsection{Architecture}
Our proposed network architecture is illustrated in Figure~\ref{fig2}, which contains
two convolutional neural networks: left CNN $f_1(\cdot;\theta_1)$ and right CNN $f_2(\cdot;\theta_2)$ with weights $(\theta_1, \theta_2)$. For given input data pairs $(x_1,x_2)$, the two outputs of left CNN and right CNN are $f_1(x_1;\theta_1)$ and $f_2(x_2;\theta_2)$ respectively. Both CNNs contain $k$ hidden layers $\{h_1, h_2, ..., h_k\}$ and
each layer of the first $k-1$ layers, contains a convolution layer, batch normalization, followed by the activation function ReLU. The $k$-th layer is a linear projection of the output of $h_{k-1}$, followed by the sigmoid activation function.
The BNN output of given input data pairs $(x_1,x_2)$ is the Euclidean distance of the two outputs of left CNN and right CNN, that is defined as
\be\label{BNN_output}
f(x_1,x_2;\theta_1,\theta_2) = \frac{1}{\sqrt{n}}||(f_1(x_1;\theta_1) -  f_2(x_2;\theta_2))||,
\ee
where $n$ is the dimension of the output of $h_k$, i.e., the common representation.
We use a predefined threshold parameter $\gamma \in (0,1)$ on  $f(x_1,x_2;\theta_1,\theta_2)$ to determine whether input data pairs $(x_1,x_2)$ have a potential relationship or not.

According to the definition of the positive sample set $S_p$ and negative sample set $S_n$, BNN is supposed to judge that the data pair in $S_p$ is relevant and the data pair in $S_n$ is not relevant. Thus the BNN output $f(x_1,x_2;\theta_1,\theta_2)$ is supposed to be close to $0$ if $(x_1,x_2) \in S_p$ and $1$ if $(x_1,x_2) \in S_n$. 
Then define the loss on positive sample set as
\beq \label{loss_pos}
&l_p(S_p;\theta_1,\theta_2)= \\
&\frac{1}{|S_p|} \sum_{(x_1,x_2) \in S_p} (f(x_1,x_2;\theta_1,\theta_2) - 0)^2,
\eeq
and the loss on negative sample set as
\beq \label{loss_neg}
&l_n(S_n;\theta_1,\theta_2)= \\
&\frac{1}{|S_n|} \sum_{(x_1,x_2) \in S_n} (f(x_1,x_2;\theta_1,\theta_2) - 1)^2, 
\eeq
where BNN can be seen as a binary classification problem.
Thus the overall loss of BNN on $S_p$ and $S_n$ is
\beq\label{loss}
&L_{bnn}(S_p, S_n;\theta_1,\theta_2)= \\
&\frac{l_p(S_p;\theta_1,\theta_2) +\alpha \cdot l_n(S_n;\theta_1,\theta_2)}{1+\alpha},
\eeq
where $\alpha$ is a parameter to balance the weights of positive samples and negative samples. Then the goal of BNN is, given $S_p$ and $S_n$, to find weights $(\theta_1^*, \theta_2^*)$ such that
\be\label{goal}
(\theta_1^*, \theta_2^*)  = {\rm argmin}_{\theta_1,\theta_2} l(S_p, S_n;\theta_1,\theta_2).
\ee

{\bf Relation with DCCA}. Here we present a discussion of the relation between BNN and DCCA~\cite{andrew2013deep}. Besides the slight difference of basic architecture, i.e. DCCA uses fully connected layers and BNN uses convolutional layers, the main difference between DCCA and BNN is the optimization objective function.
Since the motivation of DCCA is from the goal of canonical correlation analysis, the objective of DCCA is to find $(\theta_1^*, \theta_2^*)$ which make the output layers of left and right CNN maximally correlated, that is
\be\label{goal_DCCA}
(\theta_1^*, \theta_2^*)  = {\rm argmax}_{\theta_1,\theta_2}\,{\rm corr}(f_1(X_1;\theta_1), f_2(X_2;\theta_2)),
\ee
which just involves the positive samples $S_p$. As that discussed in \cite{andrew2013deep}, since the correlation objective is a function of the entire training set that does not decompose into a sum over data points, the stochastic optimization procedure cannot be directly used. Thus, DCCA proposed a L-BFGS based pre-training strategy for the stochastic method on mini-batches and achieved much better results. However, the objective function of BNN is Euclidean loss involving both positive samples and negative samples, which can be easily used to a stochastic optimization procedure that operates on mini-batch data points one at a time. Actually, the objective function of BNN can be seen as an alternative to the total correlation \eqref{goal_DCCA}, as that shown in the following theorem.

\begin{theorem}\label{thm1}
	For a group of weights $(\theta_1^*, \theta_2^*)$ obtained by \eqref{goal},	if the overall loss of BNN on $S_p$ and $S_n$ has $l(S_p, S_n;\theta^*_1,\theta^*_2)\rightarrow 0$, then the outputs of BNN has ${\rm corr}(f_1(X_1;\theta^*_1), f_2(X_2;\theta^*_2))\rightarrow n$ as $N \rightarrow \infty$.
\end{theorem}

\noindent{\bf Proof.} To make the comparison more clear, we use the similar notations as that in DCCA \cite{andrew2013deep}. Let $H_1 \in \R^{n \times N}$, $H_2 \in \R^{n \times N}$ be matrices whose columns are the outputs produced by the left CNN and right CNN on $X_1$ and $X_2$ of size $N$ with weights $(\theta_1^*, \theta_2^*)$. That is $H_1 = f_1(X_1;\theta^*_1)$, $H_2 = f_2(X_2;\theta^*_2)$.
Let $\bar H_1 = H_1 - \frac{1}{N}H_1{\bf 1}$ be the centered data matrix (resp. $\bar H_2$), and define $\hat \Sigma_{12} = \frac{1}{N-1} \bar H_1 \bar H'_2$, and $\hat \Sigma_{11} = \frac{1}{N-1} \bar H_1 \bar H'_1 + r_1 I$ for regularization constant $r_1$ (resp. $\hat \Sigma_{22}$). Here $r_1 > 0$ is chosen to make $\hat \Sigma_{11}$ positive definite. Then by the discussion of Section 2 in \cite{andrew2013deep}, the correlation, i.e., the total correlation of the top $n$ components of $H_1$ and $H_2$ is the sum of the top $n$ singular values of matrix $T = \hat \Sigma_{11}^{-1/2} \hat \Sigma_{12} \hat \Sigma_{22}^{-1/2}$, that is exactly the trace of matrix $T$:
\beq\label{corr_matrix}
{\rm corr}(H_1, H_2) = ||T||_{\rm tr} = {\rm tr}(T'T)^{1/2}.
\eeq

Now we come to prove this theorem. For a group of weights $(\theta_1^*, \theta_2^*)$ obtained by \eqref{goal}, if the overall loss of BNN on $S_p$ and $S_n$ has $l(S_p, S_n;\theta^*_1,\theta^*_2)\rightarrow 0$, by the definition in \eqref{loss}, we have
\beqn
l_p(S_p;\theta^*_1,\theta^*_2) \rightarrow 0, \quad l_n(S_n;\theta^*_1,\theta^*_2) \rightarrow 0,
\eeqn
combined with the definition in \eqref{loss_pos} and \eqref{loss_neg}, we obtain
\beqn
f(x_1,x_2;\theta^*_1,\theta^*_2) \rightarrow 0, \quad \forall\, (x_1,x_2) \in S_p, \\
f(x_1,x_2;\theta^*_1,\theta^*_2) \rightarrow 1, \quad \forall\, (x_1,x_2) \in S_n.
\eeqn
By the definition of $f(x_1,x_2;\theta^*_1,\theta^*_2)$ in \eqref{BNN_output} and $H_1 = f_1(X_1;\theta^*_1)$, $H_2 = f_2(X_2;\theta^*_2)$, the above equations indicate that for the columns of $H_1$, $h^i_1$, $i=1,...,N$ and the columns of $H_2$, $h^i_2$, $i=1,...,N$,
\beq\label{thm:1}
||h^i_1 - h^i_2||  \rightarrow 0, \quad ||h^i_1 - h^j_2||  \rightarrow \sqrt{n}, i\neq j,
\eeq
where the first equation means $H_1 \rightarrow H_2$, which yields
\beq\label{thm:2}
\bar H_1 \rightarrow \bar H_2.
\eeq
Notice that the output of each CNN is followed with sigmoid activation, then each entry of vector $h^i_1$ or $h^i_2$ is in $(0,1)$, which yields the $l$-th entry satisfies $|h^i_1(l) - h^i_2(l)| \in (0,1)$.
Combined with the second equation in \eqref{thm:1}, we obtain for $l=1,...,n$, 
\beq
|h^i_1(l) - h^j_2(l)|  \rightarrow 1, i\neq j.
\eeq
And since $h^i_1(l), h^j_2(l) \in (0,1)$, then we have either $h^i_1(l) \rightarrow 0$ or $h^i_2(l) \rightarrow 0$, which leads to $h^i_1(l)h^j_2(l) = 0$, $l=1,...,n$.
This means the inner product of vector $h^i_1$ and $h^i_2$ has $\langle h^i_1,h^j_2 \rangle \rightarrow 0$, $i\neq j$, leading to $H_2'H_1 \rightarrow D$, where $D= {\rm diag}(\langle h^1_1,h^1_2 \rangle, \langle h^2_1,h^2_2 \rangle, ..., \langle h^N_1,h^N_2 \rangle)$. Since $H_1 \rightarrow H_2$, we have $\langle h^i_1,h^i_2 \rangle \rightarrow ||h^i_1||^2$, which means $D$ is close to a positive diagonal matrix.
By the definition of centered data matrix $\bar H_1$, $\bar H_2$ and a simple calculation, as $N \rightarrow \infty$, we have
\beq\label{thm:3}
\bar H_2' \bar H_1 \rightarrow D.
\eeq

In order to simplify the discussion, suppose we have $N = Kn$ with $K\in \Z$.
Then we can redefine $\bar H_i = (\bar H_{i,1},\bar H_{i,2},...,\bar H_{i,K})$ with $\bar H_{i,j} \in \R^{n\times n}$, $i=1,2$, $j=1,...,K$. Then the diagonal matrix $D$ can also be redefined to $D = (D_1,D_2,...D_K)$ with $D_j \in \R^{n\times n}$, $j=1,...,K$.
Then the equation \eqref{thm:3} means $\bar H_{2,j}' \bar H_{1,j} \rightarrow D_j$.
Since $D$ is close to a positive diagonal matrix, combined with lemma that the left inverse of a square matrix is the right inverse, we have $\bar H_{1,j} \bar H_{2,j}' \rightarrow D_j$ and $\bar H_{1,j} \rightarrow \bar H_{2,j}$ according to \eqref{thm:2}, thus we obtain $\bar H_{2,j} \bar H_{1,j}' \rightarrow D_j$. Then we have 
$\bar H_2 \bar H_1' \rightarrow \hat D:=\sum_{j=1}^K D_j$, combined with \eqref{thm:2}, we have
\beq\label{thm:4}
\bar H_1 \bar H_1',\ \bar H_2 \bar H_2',\ \bar H_2 \bar H_1' \rightarrow \hat D,
\eeq
which means that the matrices $\hat \Sigma_{11},\hat \Sigma_{12},\hat \Sigma_{22}$ are close to a same positive diagonal matrix, leading to $T \rightarrow I$.
Combined with the definition of ${\rm corr}(H_1, H_2)$ in \eqref{corr_matrix}, and $H_1 = f_1(X_1;\theta^*_1)$, $H_2 = f_2(X_2;\theta^*_2)$, as $N \rightarrow \infty$ we have
\beqn
{\rm corr}(f_1(X_1;\theta^*_1), f_2(X_2;\theta^*_2))\rightarrow n.
\eeqn
This ends the proof.

We should remark that the above proof can be just asymptotically understood.
Actually, since we always have $n \ll N$, it's impossible to have $H_2'H_1$ be equal to a diagonal matrix. Theorem \ref{thm1} is proposed to show that the objective of BNN is asymptotically equivalent to maximizing the correlation of the outputs of two CNNs on input data pairs, i.e. two views as that named in DCCA. In conclusion, the benefits of using BNN than DCCA can be summarized as
\begin{itemize}
	\item Both for 1D signals and 2D signals, DCCA uses fully connected layers, while BNN uses convolutional layers, which makes BNN less computation cost. There are around $10^6$ learnable parameters in DCCA model, while only about $10^5$ parameters exist in our model.
	\item DCCA uses L-BFGS based pre-training strategy for a stochastic method on mini-batch training, but with no theoretical analysis to verify the effectiveness of the pre-training strategy and more cost in searching the pre-trained model. BNN proposes a novel optimization objective with asymptotic equivalence to the maximization of the total correlation, and the objective can be easily used to a stochastic optimization procedure that operates on mini-batch data points one at a time.
\end{itemize}
Actually, there is a potential relation of using L-BFGS based pre-training strategy on the architecture for DCCA. Since DCCA uses fully connected layers, the pre-training of the weights in each layer can be formulated to find a local minimum of the total squared error of a reconstructing problem, where the L-BFGS second-order optimization method can be used. Otherwise, if DCCA uses convolutional layers, it will be complicated to use L-BFGS, since L-BFGS is more suitable to optimize a matrix. 

{\bf Relation with 2WayNet}. The 2WayNet~\cite{eisenschtat2017linking} is a novel, bi-directional neural network architecture for the task of matching vectors from two data sources. The main contribution of 2WayNet is using Euclidean loss and proposing some techniques to overcome the common Euclidean regression optimization problems. In conclusion, the benefits of using BNN than 2WayNet can be summarized as
\begin{itemize}
	\item Both for 1D signals and 2D signals, 2WayNet uses fully connected layers to make the signal able to be reconstructed (but no reconstruction results are found in 2WayNet), while BNN uses convolutional layers, which makes BNN gain performance benefits when dealing with high dimensional inputs, such as datasets of large size images.
	\item 2WayNet gives the lower bound analysis of the total correlation on the Euclidean loss, but with no upper bound analysis. Instead of just using Euclidean loss, the artificial negative sample is introduced in BNN to make the objective asymptotically equivalent to the total correlation, which can be seen as another technique to handle the Euclidean regression optimization problems. 
\end{itemize}

{\bf Relation with Siamese Network}. Siamese Network ~\cite{chopra2005learning} is a classical neural network architecture, which learns the similarity between two inputs. It consists of two identical neural networks, each taking one of the two inputs. The similarity between them is then calculated by feeding the last layers of the two networks into a contrastive loss function. BNN has a similar positive/negative-sample based optimization objective with Siamese Network, however there are three fundamental differences between them.
\begin{itemize}
	\item The main task of Siamese Network is pair matching problem, while BNN focuses on common representation learning which can be used in variety tasks.
	\item The loss function of Siamese Network is a summation of product between label and a
	Euclidean-based pre-designed loss, while the loss function of BNN is a Euclidean-based binary classification loss.
	\item Siamese Network shares weights for right and left networks, while BNN has two totally independent networks which enable it to solve multimodal problems naturally.
\end{itemize}

\subsection{Training process}
Once the BNN model, i.e.  $\{h_1, h_2, ..., h_k\}$ of left CNN and right CNN, is defined, we can use a stochastic optimization procedure, e.g. SGD with mini-batch to train the weights given the training samples.  Here we train the left CNN and the right CNN alternately in a single iteration, i.e. one mini-batch.
As that discussed before, the negative samples used are randomly selected from $S_n$ according to a predefined parameter $\xi$, named NP ratio, which denotes the ratio of the size of the negative samples on that of positive samples. See Algorithm \ref{alg} for the details of the training process of BNN. 

Actually, using randomly selected negative samples is a trade-off strategy. In the ideal condition, we hope using all negative samples in the training, which is impossible in practice. We compared the performance of the proposed negative sample generating strategy and the ideal one (training with all negative samples) based on a small sub-dataset. The difference between the results of them is negligible. 

\begin{algorithm}[!ht]
	\caption{Training process of BNN}\label{alg}
	\begin{algorithmic}[1]
		\STATE Construct positive samples $S_p = \{x_1^i, x_2^i\}$;
		\STATE Define balance parameter $\alpha$, learning rate $\eta$ and NP ratio $\xi$;
		\STATE Randomly initialize weights $\theta_1, \theta_2$;
		\FOR{Each epoch}
		\STATE Randomly construct negative samples: \\ $S_n = \{x_1^i, x_2^j\}, i\neq j$ with size $N\xi$;
		\FOR{mini-batch $S^i_p \subset S_p,  S^i_n \subset S_n$}
		\STATE Update left CNN with \\ $\theta_1 \gets \theta_1 - \eta \frac{\partial l_{bnn}(S^i_p, S^i_n;\theta_1,\theta_2)}{\partial \theta_1}$;
		\STATE Update right CNN with \\ $\theta_2 \gets \theta_2 - \eta \frac{\partial l_{bnn}(S^i_p, S^i_n;\theta_1,\theta_2)}{\partial \theta_2}$;
		\ENDFOR
		\ENDFOR
	\end{algorithmic}
\end{algorithm}

\section{Experiments}
In this section, three experiments are presented, including pair matching test, Canonical Correlation Analysis and transfer learning across views. We perform our experiments based on two datasets commonly used in the recent literature for CCA test: MNIST~\cite{lecun-98} half matching and X-Ray Microbeam Speech data (XRMB)~\cite{westbury1994x}. As for the model we use in our experiment, both the left and right networks have structures of 3 successive convolution layers with 10 filters ($3\times3$ filters for MNIST, $1\times3$ filters for XRMB) and batch normalization. Each output of the last convolution layers of the two networks connects to a fully connected layer, linearizing into a $n$-dimensional vector. 

Actually, CNN is an optional choice in BNN. However, the experiments demonstrate that using lightweight CNN is a better choice since it can achieve the similar performances as that of using full-connected layers (For MNIST dataset, the corresponding accuracy, precision, recall of pair matching and CCA result is 92.10, 84.61, 93.29, 49.23). Here we ignore the detailed discussion of comparison.  

\subsection{Data Description}

\subsubsection*{MNIST half matching}
MNIST handwritten digits image dataset~\cite{lecun-98} is commonly used for training various image processing systems. The database contains 60,000 training images and 10,000 testing images of $28\times28$ pixels. In our experiments, each image is cut into two $28\times14$ pixel halves vertically. Two strategies are used to construct positive and negative samples: 1) For matching test and CCA examination, we only consider two halves of the same images as positive samples. 2) For transfer learning across views, all pairs of halves from the same digits (with the same labels) are set as positive samples.

\subsubsection*{X-Ray Microbeam Speech data (XRMB)}
In XRMB dataset~\cite{westbury1994x}, two sets of data are adopted: simultaneous acoustic and articulatory recordings. The acoustic views are MFCCs with 273-dimensional vector per frame, while the articulatory data is represented as a 112-dimensional vector. In our experiment, 60,000 random samples are used for training and 10,000 for testing. Simultaneous acoustic and articulatory recording pairs are considered as positive samples.

\subsection{Pair Matching Across Two Views}
Among the existing works mentioned before, this task has not been explicitly implemented. Here we will only show our own results. 
On the testing data, we generate 10,000 positive pairs and 20,000 negative pairs. The output dimension $n$ of both networks is set to 50 for MNIST dataset, 112 for XRMB dataset, and the judging threshold $\gamma$ is set to 0.5. To make the results more detailed, we also show the precision, recall, and F1 score in addition to the accuracy measurement.

\begin{table}[htb]
	\caption{Performance of pair matching test across two views learned by BNN on MNIST dataset (\%).}
	\label{Prediction accuracy}
	\centering
	\begin{tabular}{ccc}
		\toprule
		  & MNIST   & XRMB   \\
		\midrule
		Accuracy   & 95.63  &   87.77 \\
		Precision & 88.71  & 75.64 \\
		Recall & 99.58 & 93.39 \\
		F1 Score & 93.83 &   83.58 \\
		\bottomrule
	\end{tabular}
\end{table}

Table~\ref{Prediction accuracy} suggests that BNN has a outstanding performance
on pair matching test, especially the value of recall rate, which proves that BNN is able to learning exact features for accurately predict the relationship
of two views as we expected.

\begin{figure}[htb]
	\centering
	\includegraphics[width=0.46\textwidth]{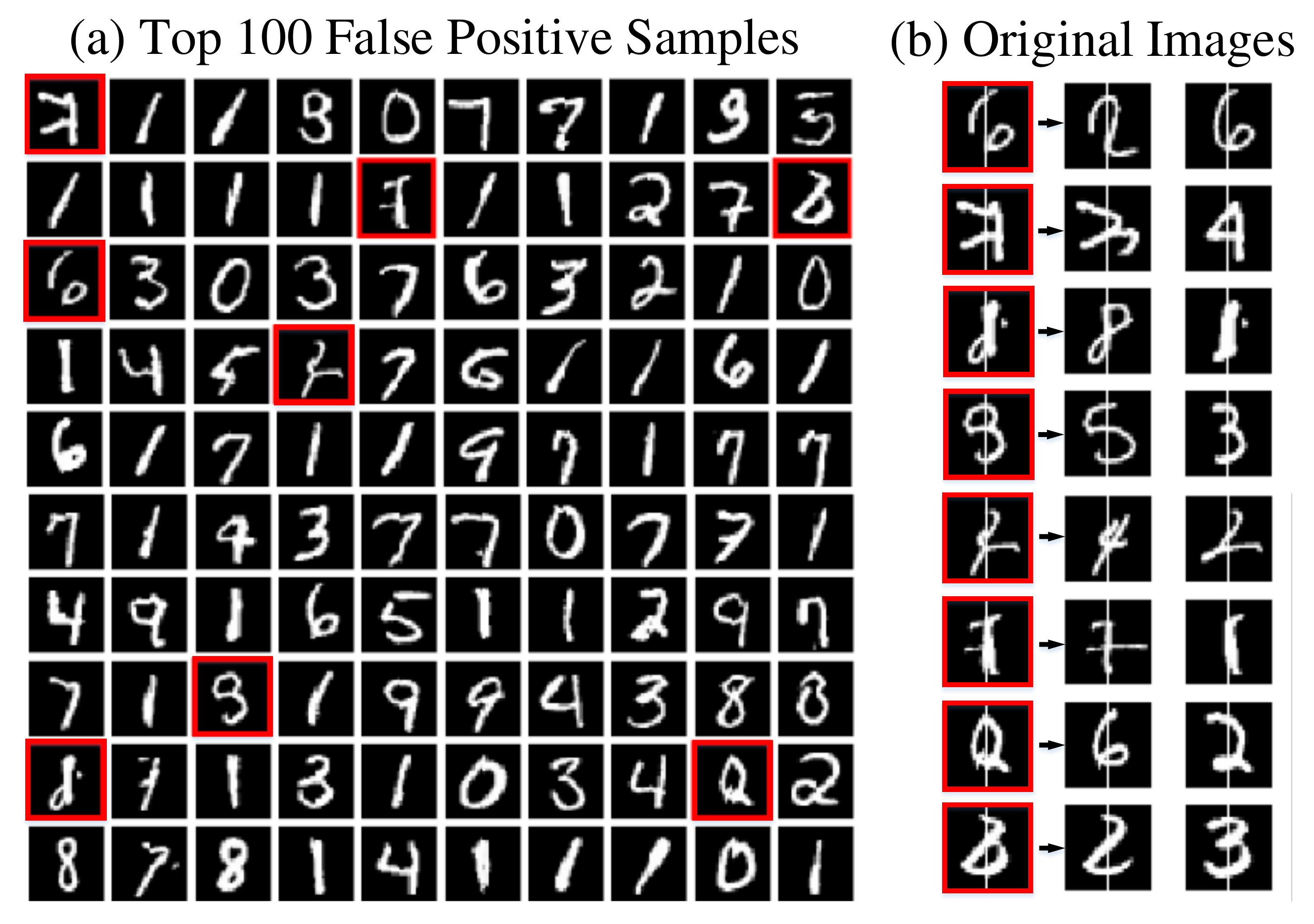}
	\caption{(a) Top 100 false positive samples. Samples outlined in red are obviously negative for humans;
		(b) The original two images of those negative pairs outlined in (a).}
	\label{fp}
\end{figure}

Figure~\ref{fp} (a) shows top 100 false positive samples on MNIST dataset. Most of those pairs are joined smoothly so that even humans will regard them as positive samples visually. Though some of them are obviously negative samples, their original images are illegible (See Figure~\ref{fp} (b)). In this case, the results can be considered as reasonable and acceptable.

\subsection{Correlation between representations of two views}
In this section, we will present the CCA result of BNN and compare it with the models we
mentioned above. We follow Andrew's Method~\cite{andrew2013deep} calculating the total correlation captured in the $n$=50 (MNIST) or $n$=112 (XRMB) dimensions of the common representations learned by the BNN. The structure of networks is the same as the one in the previous experiment. However, the testing data only contains positive samples, the same as all the experiments in the other works. The results are reported in Table~\ref{CCA}. All CCA results of comparative models are stated in the literature of 2WayNet~\cite{eisenschtat2017linking}.

\begin{table}[htb]
	\caption{
		Total correlation of the common representations
		learned by different models on MNIST and XRMB datasets.}
	\label{CCA}
	\centering
		\begin{tabular}{ccc}
			\toprule
			Methods  & MNIST & XRMB  \\
			\midrule
			CCA  & 28  & 16.9 \\
			DCCA  &  39.7  &  92.9 \\
			RCCA &  44.5 & 104.5 \\
			DCCAE & 25.34 & 41.47 \\
			CorrNet & 48.07  & 95.01 \\
			2WayNet & 49.15 & 110.18 \\
			\textbf{BNN} & \textbf{49.32} & \textbf{110.65} \\
			\midrule
			Upper Bound & 50 & 112\\
			\bottomrule
		\end{tabular}
\end{table}

The reported values in Table~\ref{CCA} are the sum of the correlations captured in the 50 (MNIST) or 112 (XRMB) dimensions of the learned representations of the two views. The total correlations learned by BNN are closer to the maximal value of 50 (MNIST) and 112 (XRMB), which are clearly better than the other models. Since the CCA results of 2WayNet are already very close to the maximum, any slight improvement can be considered significant.

\subsection{Transfer Learning Across Views}
In this section, all tests are implemented based on MNIST dataset. The experiment shows that the 50-dimension common representation learned in BNN from the half views of the images can be trained to predict the digits of the other half views. Linear SVM classifier provided by Scikit-learn~\cite{pedregosa2011scikit} is used in our experiment and the common representaion data used to trian the SVM classifier is inferenced by a well-trained and fixed BNN model. For each model list in Figure~\ref{Transfer learning}, we report 5-fold cross-validation accuracy on 10,000 images in the MNIST test dataset. Two sets of tests are reported: (i) training on the left views and testing on the right views, (ii) training on the right views and testing on the left views. All transfer learning results of comparative models are stated in CorrNet~\cite{chandar2016correlational}.

\begin{table}[htb]
	\caption{Transfer learning (Left to Right, Right to Left) accuracy using the representations learned using different
		models on the MNIST dataset.}
	\label{Transfer learning}
	\centering
	\begin{tabular}{ccc}
		\toprule
		Methods &  L. to R.  &  R. to L.  \\
		\midrule
		CCA & 65.73  & 65.44    \\
		DCCA   &  68.10 & 75.71    \\
		MAE & 64.14 &  68.88   \\
		CorrNet  &  77.05  &  78.81   \\
		\textbf{BNN} & \textbf{90.21} & \textbf{91.38} \\
		\midrule
		\textbf{Single View}  & \textbf{93.17} & \textbf{91.52}   \\
		\bottomrule
	\end{tabular}
\end{table}

In Table~\ref{Transfer learning}, single view corresponds to the classifier trained and tested on the same halves of images. 
The value of single view can represent the effectiveness of the learned common representation. 
Compared with the singe view results of CorrNet~\cite{chandar2016correlational}, which are only 81.62 and 80.06 for left and right single views, the single view results of BNN show that the learned common representations with only half digits are good enough for classification. 
In addition, the single view result is also the upper bound of our transfer learning performance. For BNN, the result of transfer learning accuracy is highly close to the accuracy of ideal single view case, indicating that the common representations captured by BNN gains excellent ability of transfer learning. More importantly, it performs significantly better than all the other models as well. 
The high accuracy of transfer learning and single view on the representation indicates that the common representation found by BNN is much closer to the ideal common feature space for the classification task. Compared with the limited improvement on the task of total correlation captured in Table~\ref{CCA}, the huge improvement on transfer learning demonstrates the efficiency of using different positive samples task by task, since the training samples used in comparison methods are consistent across tasks.

\subsection{Reconstruction Across Views}
\begin{figure}[htb]
	\begin{center}
		\includegraphics[width=0.46\textwidth]{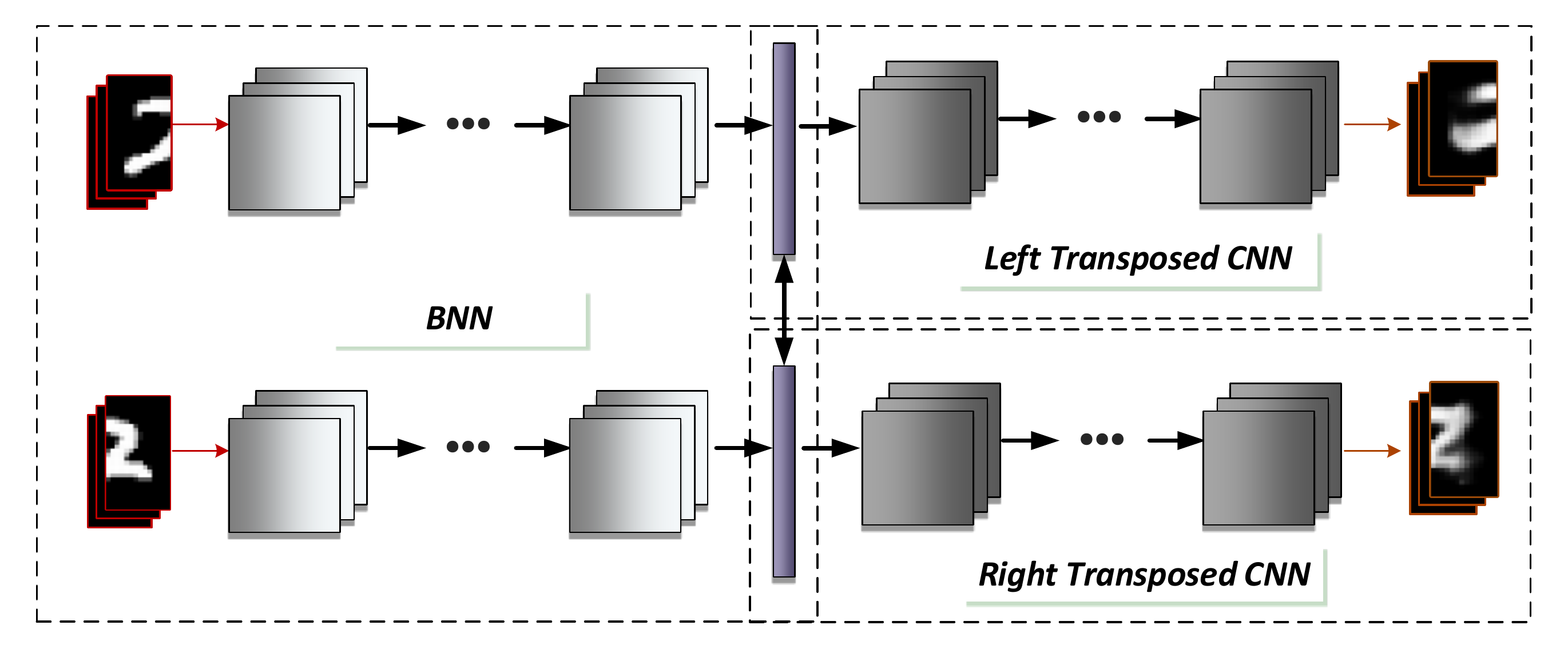}
	\end{center}
	\caption{A schematic of bridge neural network with reconstruction structure, which employs a standard BNN~\eqref{fig2} and two transposed convolutional neural networks.}\label{reconmodel}
\end{figure}

Firstly, we present the reconstruction architecture using BNN. 
Although MAE structure is unnecessary to learn common representations in our algorithm, view reconstruction can be achieved by adding two transposed convolution networks following the common representation layers of BNN. 
The architecture is illustrated in Figure~\ref{reconmodel} which contains a standard bridge neural network and two transposed convolutional neural networks. For given input data pairs $(x_1,x_2)$, the two outputs of the left CNN and right CNN are $f_1(x_1;\theta_1)$ and $f_2(x_2;\theta_2)$ respectively. For the given common representations $(z_1,z_2)$ of the two views, the reconstruction of them are the outputs of the two transposed convolutional neural networks, which are $f_1'(z_1;\theta_1')$ and $f_2'(z_2;\theta_2')$ respectively. 

To minimize the reconstruction error, new loss functions are introduced beside the loss of BNN~\eqref{loss}. 
For self-reconstructions, we have the outputs of the two transposed convolutional neural networks as 
$i=1,2$,  
\beq
x_{i\_sel}'= f_i'(f_i(x_i;\theta_i);\theta_i'). 
\eeq
Then we define the loss of self-reconstruction on $S_p$ as 
\beq
&l_{self}(S_p;\theta_1,\theta_2,\theta_1',\theta_2') =  \\
&\frac{1}{|S_p|} \sum_{(x_1,x_2) \in S_p}{(||x_{1\_sel}'-x_1||+||x_{2\_sel}'-x_2||)}.
\eeq

For cross-reconstructions, we have two outputs as
\beq
x_{1\_cro}'&= f_1'(f_2(x_2;\theta_2);\theta_1'), \\
x_{2\_cro}'&= f_2'(f_1(x_1;\theta_1);\theta_2').
\eeq
Then we define the loss of cross-reconstruction on $S_p$ as 
\beq
&l_{cross}(S_p;\theta_1,\theta_2,\theta_1',\theta_2') = \\
&\frac{1}{|S_p|} \sum_{(x_1,x_2) \in S_p}{(||x_{1\_cro}'-x_1||+||x_{2\_cro}'-x_2||)}.
\eeq
Thus the overall loss of BNN with reconstruction is
\be
l_{total}(S_p, S_n;\theta_1,\theta_2,\theta_1',\theta_2')=l_{bnn}+l_{self}+ l_{cross}. 
\ee
In the training process, i.e. Algorithm \ref{alg}, we replace the loss $l_{bnn}$ with $l_{total}$ and update the weights in transposed convolutional neural networks $\theta_i'$ together with $\theta_i$, $i=1,2$. 

\begin{figure}[htb]
	\centering
	\includegraphics[width=0.47\textwidth]{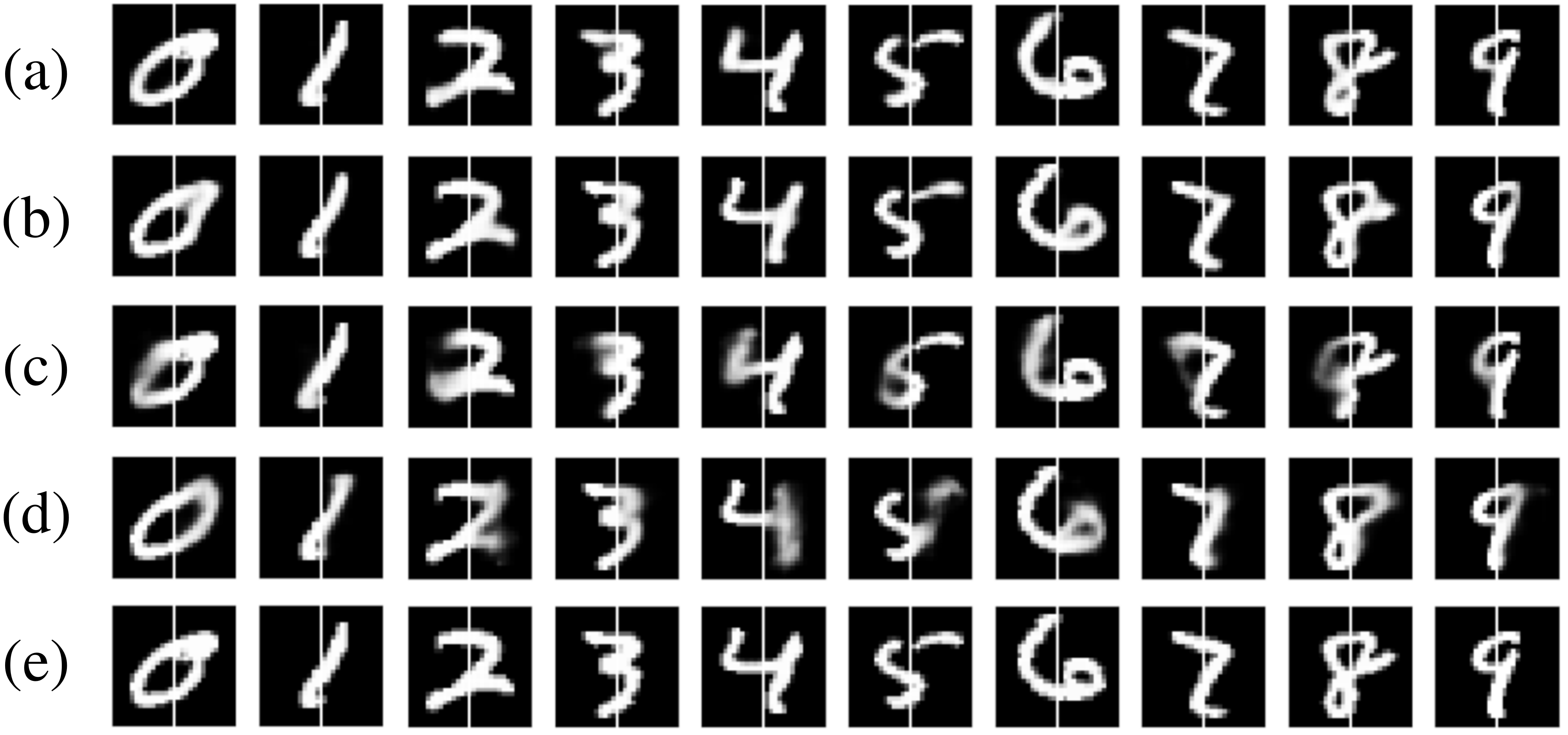}
	\caption{Reconstruction results of BNN for MNIST. (a) Left view self-reconstruction; (b) Right view self-reconstruction; (c) Right to left cross-view reconstruction; (d) Left to right cross-view reconstruction; (e) Original images of two views. }
	\label{fig4}
\end{figure}

In this experiment, view reconstruction is implemented on the MNIST dataset. The structure of BNN is the same as previous. As for the two transposed convolutional networks, in each of them, the 50-dimension common representation layer connects to a $50\times64$ fully connected layer and then reshaped into a $4\times2\times8$ matrix. It is followed by 3 successive transposed convolution layers with $8/16/32$ filters respectively ($3\times3$ kernel), outputting a $28\times14\times32$ matrix. At last, one $3\times3$ filter convolutional layer will project the matrix into a $28\times14\times1$ output image. 

Figure~\ref{fig4} shows the output images of self-reconstruction and cross-reconstruction both on left and right views of a few samples. The results are more clear and sharp both in self-reconstruction and cross-reconstruction compared with the visual results of CorrNet~\cite{chandar2016correlational}. The visually satisfying, which indicates that reconstruction can also be an optional function for Bridge Neural Network.

\section{Conclusion}
This paper proposes bridge neural network (BNN) with lightweight convolution layers to dig the potential relationship of two given data sources, that can be specified according to a computer vision task. The training objective with the artificial negative samples is introduced and it's asymptotically equivalent to maximizing the total correlation of the two data sources in theory. The experiments on the tasks, including pair matching, canonical correlation analysis, transfer learning and reconstruction demonstrate the state-of-the-art performance of BNN. 

\section{Acknowledgements}
This work was supported by the Innovation
Foundation of Qian Xuesen Laboratory of Space Technology and the National Natural Science Foundation of China (61702520).

\bibliography{AAAI-XuY.4068}

\begin{thebibliography}{}

\bibitem[\protect\citeauthoryear{Akaho}{2006}]{akaho2006kernel}
Akaho, S.
\newblock 2006.
\newblock A kernel method for canonical correlation analysis.
\newblock {\em arXiv preprint cs/0609071}.

\bibitem[\protect\citeauthoryear{Andrew \bgroup et al\mbox.\egroup
  }{2013}]{andrew2013deep}
Andrew, G.; Arora, R.; Bilmes, J.; and Livescu, K.
\newblock 2013.
\newblock Deep canonical correlation analysis.
\newblock In {\em International Conference on Machine Learning},  1247--1255.

\bibitem[\protect\citeauthoryear{Arora and Livescu}{2013}]{arora2013multi}
Arora, R., and Livescu, K.
\newblock 2013.
\newblock Multi-view cca-based acoustic features for phonetic recognition
  across speakers and domains.
\newblock In {\em Acoustics, Speech and Signal Processing (ICASSP), 2013 IEEE
  International Conference on},  7135--7139.
\newblock IEEE.

\bibitem[\protect\citeauthoryear{Bach and Jordan}{2002}]{bach2002kernel}
Bach, F.~R., and Jordan, M.~I.
\newblock 2002.
\newblock Kernel independent component analysis.
\newblock {\em Journal of machine learning research} 3(Jul):1--48.

\bibitem[\protect\citeauthoryear{Chandar \bgroup et al\mbox.\egroup
  }{2016}]{chandar2016correlational}
Chandar, S.; Khapra, M.~M.; Larochelle, H.; and Ravindran, B.
\newblock 2016.
\newblock Correlational neural networks.
\newblock {\em Neural computation} 28(2):257--285.

\bibitem[\protect\citeauthoryear{Chopra, Hadsell, and
  LeCun}{2005}]{chopra2005learning}
Chopra, S.; Hadsell, R.; and LeCun, Y.
\newblock 2005.
\newblock Learning a similarity metric discriminatively, with application to
  face verification.
\newblock In {\em Computer Vision and Pattern Recognition, 2005. CVPR 2005.
  IEEE Computer Society Conference on}, volume~1,  539--546.
\newblock IEEE.

\bibitem[\protect\citeauthoryear{Dhillon, Foster, and
  Ungar}{2011}]{dhillon2011multi}
Dhillon, P.; Foster, D.~P.; and Ungar, L.~H.
\newblock 2011.
\newblock Multi-view learning of word embeddings via cca.
\newblock In {\em Advances in neural information processing systems},
  199--207.

\bibitem[\protect\citeauthoryear{Eisenschtat and
  Wolf}{2017}]{eisenschtat2017linking}
Eisenschtat, A., and Wolf, L.
\newblock 2017.
\newblock Linking image and text with 2-way nets.
\newblock {\em arXiv preprint}.

\bibitem[\protect\citeauthoryear{Hardoon \bgroup et al\mbox.\egroup
  }{2007}]{hardoon2007unsupervised}
Hardoon, D.~R.; Mourao-Miranda, J.; Brammer, M.; and Shawe-Taylor, J.
\newblock 2007.
\newblock Unsupervised analysis of fmri data using kernel canonical
  correlation.
\newblock {\em NeuroImage} 37(4):1250--1259.

\bibitem[\protect\citeauthoryear{Hardoon, Szedmak, and
  Shawe-Taylor}{2004}]{hardoon2004canonical}
Hardoon, D.~R.; Szedmak, S.; and Shawe-Taylor, J.
\newblock 2004.
\newblock Canonical correlation analysis: An overview with application to
  learning methods.
\newblock {\em Neural computation} 16(12):2639--2664.

\bibitem[\protect\citeauthoryear{Hotelling}{1936}]{hotelling1936relations}
Hotelling, H.
\newblock 1936.
\newblock Relations between two sets of variates.
\newblock {\em Biometrika} 28(3/4):321--377.

\bibitem[\protect\citeauthoryear{Kim, Wong, and Cipolla}{2007}]{kim2007tensor}
Kim, T.-K.; Wong, S.-F.; and Cipolla, R.
\newblock 2007.
\newblock Tensor canonical correlation analysis for action classification.
\newblock In {\em Computer Vision and Pattern Recognition, 2007. CVPR'07. IEEE
  Conference on},  1--8.
\newblock IEEE.

\bibitem[\protect\citeauthoryear{LeCun \bgroup et al\mbox.\egroup
  }{1998}]{lecun-98}
LeCun, Y.; Bottou, L.; Bengio, Y.; and Haffner, P.
\newblock 1998.
\newblock Gradient-based learning applied to document recognition.
\newblock {\em Proceedings of the IEEE} 86(11):2278--2324.

\bibitem[\protect\citeauthoryear{Melzer, Reiter, and
  Bischof}{2001}]{melzer2001nonlinear}
Melzer, T.; Reiter, M.; and Bischof, H.
\newblock 2001.
\newblock Nonlinear feature extraction using generalized canonical correlation
  analysis.
\newblock In {\em International Conference on Artificial Neural Networks},
  353--360.
\newblock Springer.

\bibitem[\protect\citeauthoryear{Michaeli, Wang, and
  Livescu}{2016}]{michaeli2016nonparametric}
Michaeli, T.; Wang, W.; and Livescu, K.
\newblock 2016.
\newblock Nonparametric canonical correlation analysis.
\newblock In {\em International Conference on Machine Learning},  1967--1976.

\bibitem[\protect\citeauthoryear{Mineiro and
  Karampatziakis}{2014}]{mineiro2014randomized}
Mineiro, P., and Karampatziakis, N.
\newblock 2014.
\newblock A randomized algorithm for cca.
\newblock {\em arXiv preprint arXiv:1411.3409}.

\bibitem[\protect\citeauthoryear{Ngiam \bgroup et al\mbox.\egroup
  }{2011}]{ngiam2011multimodal}
Ngiam, J.; Khosla, A.; Kim, M.; Nam, J.; Lee, H.; and Ng, A.~Y.
\newblock 2011.
\newblock Multimodal deep learning.
\newblock In {\em Proceedings of the 28th international conference on machine
  learning (ICML-11)},  689--696.

\bibitem[\protect\citeauthoryear{Pedregosa \bgroup et al\mbox.\egroup
  }{2011}]{pedregosa2011scikit}
Pedregosa, F.; Varoquaux, G.; Gramfort, A.; Michel, V.; Thirion, B.; Grisel,
  O.; Blondel, M.; Prettenhofer, P.; Weiss, R.; Dubourg, V.; et~al.
\newblock 2011.
\newblock Scikit-learn: Machine learning in python.
\newblock {\em Journal of machine learning research} 12(Oct):2825--2830.

\bibitem[\protect\citeauthoryear{Vinod}{1976}]{vinod1976canonical}
Vinod, H.~D.
\newblock 1976.
\newblock Canonical ridge and econometrics of joint production.
\newblock {\em Journal of econometrics} 4(2):147--166.

\bibitem[\protect\citeauthoryear{Wang \bgroup et al\mbox.\egroup
  }{2015}]{wang2015deep}
Wang, W.; Arora, R.; Livescu, K.; and Bilmes, J.
\newblock 2015.
\newblock On deep multi-view representation learning.
\newblock In {\em International Conference on Machine Learning},  1083--1092.

\bibitem[\protect\citeauthoryear{Wang, Li, and
  Lazebnik}{2016}]{wang2016learning}
Wang, L.; Li, Y.; and Lazebnik, S.
\newblock 2016.
\newblock Learning deep structure-preserving image-text embeddings.
\newblock In {\em Proceedings of the IEEE conference on computer vision and
  pattern recognition},  5005--5013.

\bibitem[\protect\citeauthoryear{Westbury}{1994}]{westbury1994x}
Westbury, J.
\newblock 1994.
\newblock X-ray microbeam speech production database user’s handbook:
  Madison.
\newblock {\em WI: Waisman Center, University of Wisconsin}.

\bibitem[\protect\citeauthoryear{Xu, Tao, and Xu}{2013}]{Xu2013A}
Xu, C.; Tao, D.; and Xu, C.
\newblock 2013.
\newblock A survey on multi-view learning.
\newblock {\em Computer Science}.

\bibitem[\protect\citeauthoryear{Yan and Mikolajczyk}{2015}]{yan2015deep}
Yan, F., and Mikolajczyk, K.
\newblock 2015.
\newblock Deep correlation for matching images and text.
\newblock In {\em Computer Vision and Pattern Recognition (CVPR), 2015 IEEE
  Conference on},  3441--3450.
\newblock IEEE.

\bibitem[\protect\citeauthoryear{Zhao \bgroup et al\mbox.\egroup
  }{2017}]{Zhao2017Multi}
Zhao, J.; Xie, X.; Xu, X.; and Sun, S.
\newblock 2017.
\newblock Multi-view learning overview: Recent progress and new challenges.
\newblock {\em Information Fusion} 38:43--54.

\end{thebibliography}
\bibliographystyle{aaai}

\end{document}